\documentclass{article}
\usepackage{amsmath,epsfig}
\usepackage[preprint]{spconfa4}
\usepackage[inline]{enumitem}
\usepackage{hhline}
\usepackage{amssymb}
\usepackage{xcolor}
\usepackage{subcaption}
\usepackage{microtype}
\let\OLDthebibliography\thebibliography
\renewcommand\thebibliography[1]{
  \OLDthebibliography{#1}
  \setlength{\parskip}{0pt}
  \setlength{\itemsep}{0pt plus 0.3ex}
}

\begin{document}\sloppy

\def\x{{\mathbf x}}
\def\L{{\cal L}}

\title{Multi-object tracking with \\tracked object bounding box association}
%
\name{Nanyang~Yang, Yi~Wang and Lap-Pui~Chau}
\address{School of Electrical and Electronic Engineering, Nanyang Technological University, Singapore \\
yang0526@e.ntu.edu.sg, wang1241@e.ntu.edu.sg, elpchau@ntu.edu.sg}

\maketitle

\begin{abstract}
The CenterTrack tracking algorithm achieves state-of-the-art tracking performance using a simple detection model and single-frame spatial offsets to localize objects and predict their associations in a single network. However, this joint detection and tracking method still suffers from high identity switches due to the inferior association method. 
To reduce the high number of identity switches and improve the tracking accuracy, 
in this paper, we propose to incorporate a simple tracked object bounding box and overlapping prediction based on the current frame onto the CenterTrack algorithm. Specifically, we propose an Intersection over Union (IOU) distance cost matrix in the association step instead of simple point displacement distance. We evaluate our proposed tracker on the MOT17 test dataset, showing that our proposed method can reduce identity switches significantly by 22.6\% and obtain a notable improvement of 1.5\% in IDF1 compared to the original CenterTrack's under the same tracklet lifetime. The source code is released at https://github.com/Nanyangny/CenterTrack-IOU.

\end{abstract}
\section{Introduction}
\label{sec:intro}

Multi-object tracking (MOT) is a popular topic in computer vision due to its wide application in areas such as transportation and elderly care. Recent progress on joint detection and tracking technique has drawn much research attention in MOT problems. MOT is a task to estimate trajectories for objects of interest through space and time \cite{introspace_time}. The rapid development of deep learning has advanced the research on MOT.

MOT is often addressed by the tracking-by-detection paradigm which consists of two parts \cite{zhang2020fair}. First, an object detection algorithm that outputs detection results in the form of bounding boxes location in every frame; then an association algorithm is used to link up the newly detected objects with the existing tracks based on spatial information or extracted re-identification (re-ID) features, or both. Most of the existing MOT solvers use two separate models to perform the two steps respectively. Although there has been significant development in object detection \cite{fasterRcnn,zhou2019objects,redmon2018yolov3} and re-ID \cite{rtpeople,zheng2017person} separately to enhance the overall tracking performance, those methods hardly achieve real-time inference speed due to the slow and complex association methods \cite{hu2019joint,People_Tracking,7410891} and separate learned models without shared features. 

Recent research in simultaneous detection and tracking method \cite{tracking_bw,dtc_ctd,zhou2020tracking} provides another viable research direction for MOT tasks. Under this approach, existing detectors are converted into trackers and both tasks are combined in the same framework. Two tasks now share the same set of low-level features, therefore no need for re-computation. 

\begin{figure}[t]
    \centering
    \includegraphics[width=1\linewidth]{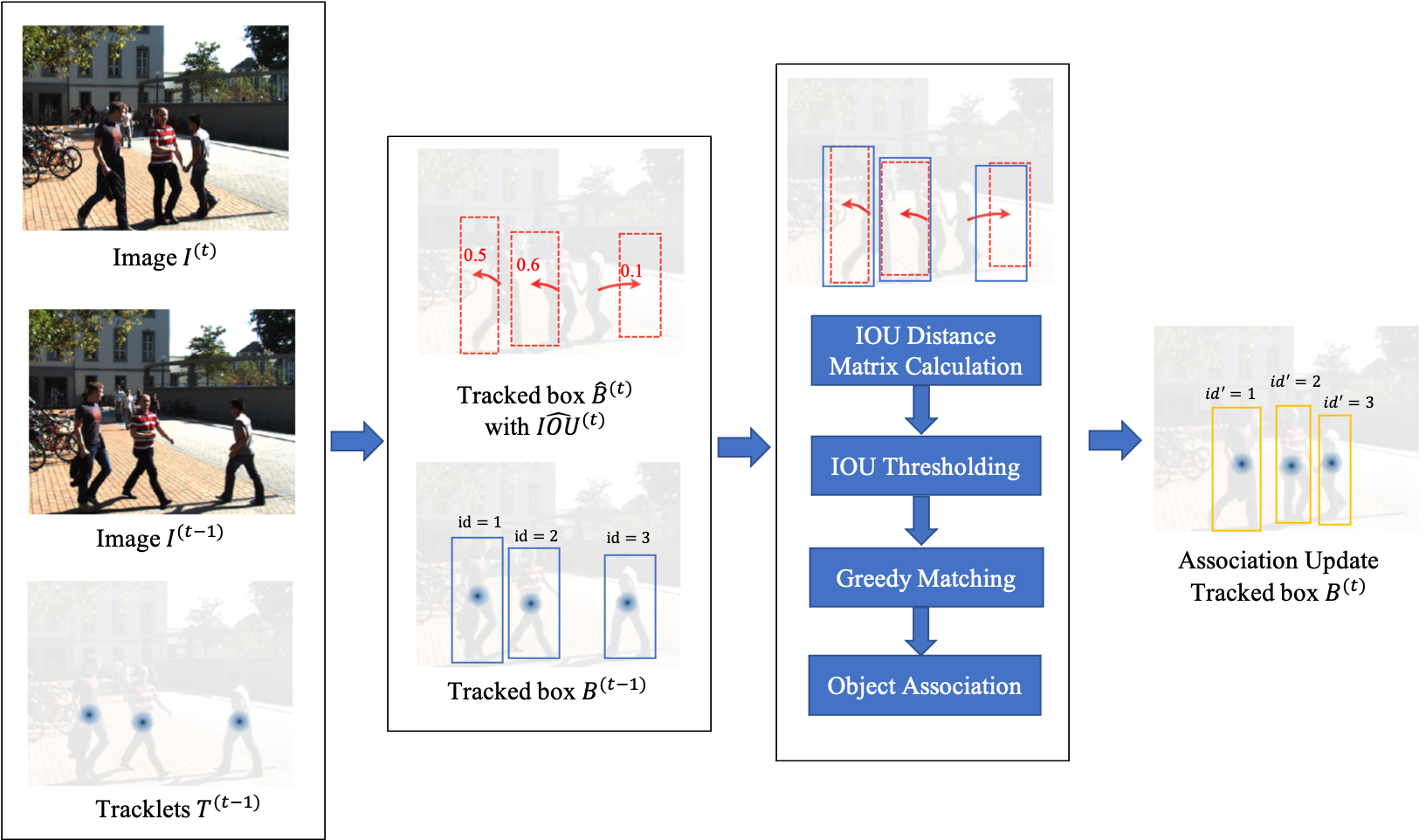}
    \caption{Overview of our proposed method: A network is configured to predict tracked bounding box and IOU based on the original CenterTrack model. IOU distance cost matrix is calculated between the tracked object bounding boxes $\hat{B}^{(t-1)}$ (in blue) at the previous frame and tracked object bounding boxes prediction $\hat{B}^{(t)}$ (in dotted red) from the current frame, followed by IOU filtering. Finally, a simple greedy matching algorithm is used to associate objects in the current frame.}
    \label{fig:idea}
\end{figure}

CenterTrack \cite{zhou2020tracking}, one of the state-of-the-art trackers, adopts the idea of simultaneous detection and tracking methods with point-based detection. In CenterTrack, each object is represented by the center point of its bounding box. This center point is tracked through time. Objects in a frame are represented by a heatmap of points. CenterTrack takes in heatmaps of two consecutive frames and trains the model to output an offset vector from the current object center to its center in the previous frame. A simple greedy matching is performed using the distance between the predicted offset and detected center point in the previous frame to associate object identities. This tracking-conditioned detection framework replaces the need for a motion model \cite{zhou2020tracking}, which reduces the need for extra computation. 
 However, CenterTrack relies on center displacement offset to associate objects in adjacent frames only, which is not enough to provide robust association ability especially when occlusions occur. Additionally, long-range tracklet association is not explored in \cite{zhou2020tracking}. Therefore, in this paper, we propose to integrate the prediction of the tracked object bounding box to the existing CenterTrack tracking model, which enables robust distance cost matrix calculation based on both center displacement and tracked object bounding box prediction to associate objects through long-range tracklets, shown in Figure \ref{fig:idea}. Our main contributions in this project are:
\begin{itemize}
    \item Incorporate tracked object bounding box prediction to CenterTrack using robust cost matrix calculation in object association.
    \item Evaluate the proposed method on MOT17 dataset to obtain a significant reduction in identity switches (IDs) and notable improvements in accuracy score only with additional two output branches.
\end{itemize}

This paper is organized as follows: In Section 2 and 3, the detailed methodology of the baseline baseline CenterTrack model and proposed tracking method are introduced. The experiment details and results are described in Section 4. We conclude this paper in Section 5.

\section{Baseline CenterTrack Algorithm}
CenterTrack takes three inputs, the current frame, the previous frame, and a heatmap rendered from tracked object centers. Then, the model outputs a center detection heatmap for the current frame, the bounding box size map, and a center offset map. 

\textbf{Heatmap Representation}
CenterTrack is built on CenterNet \cite{zhou2019objects}. Objects are presented by the center point of their bounding boxes. For each center \(\textbf{p}\) of class \(C\) in each frame, it is rendered as a Gaussian-shaped peak into a heatmap \( Y \in [0,1]^{\frac{W}{R}\times\frac{H}{R}\ \times C} = R(\{\textbf{p}_0,\textbf{p}_1,...\})\), the rendering function at position \(\textbf{q} \in \mathbb{R}^2\) is defined as:
\begin{equation}
    R_{q}(\{\textbf{p}_{0},\textbf{p}_{1},...\}) = \max_{i} \exp\Big(-\frac{(\textbf{p}_{i} - \textbf{q})^{2}}{2\sigma_{i}^{2}}\Big)
    \label{eqt:heatmap}
\end{equation}
where the Gaussian kernel $\sigma_{i}$ is a function of the object size \cite{law2019cornernet}.

\textbf{Tracking-conditioned Detection}
In CenterTrack, two frames are passed to the network: the current frame \(I^{(t)}\in\mathbb{R}^{W\times H\times3}\) and the prior frame \(I^{(t-1)}\in\mathbb{R}^{W\times H\times3}\). This allows the model to estimate the change between the frames and potentially reason about the occluded objects at time $t$ from visual information at time $t-1$. To help the model detect the objects in the current frame, a heatmap rendered from prior detections \( \hat{Y} \in [0,1]^{\frac{W}{R}\times\frac{H}{R}\times C} \) and size map \( \hat{S}\in \mathbb{R}^{\frac{W}{R}\times \frac{H}{R} \times 2}\), are used to feed the model as well, where $R$ is a downsampling factor and C is the number of classes. To reduce false positive detections, local maxima (peaks) in a $3 \times 3$ region are used and only peaks with a confidence score greater than a threshold $\tau$ are rendered. The object sizes are extracted from the size map to calculate the objects' bounding boxes.

\textbf{Object Association} CenterTrack predicts a center displacement as two output channels \(\hat{D}^{(t)} \in \mathbb{R}^{\frac{W}{R} \times \frac{H}{R}\times2}\). For each detected object at location $\hat{\textbf{p}}$, the predicted $\hat{D}_{\hat{p}^{(t)}}^{(t)}$ shows the difference of object center in the current frame $\hat{p}^{(t)}$ and the previous frame $\hat{p}^{(t-1)}$, $\hat{D}_{\hat{p}^{(t)}}^{(t)} = \hat{p}^{(t)} - \hat{p}^{(t-1)}$. With this center offset prediction, the object center location in the previous frame can be easily tracked. For each detection at $\hat{p}$, a simple greedy matching algorithm is used to associate it with the closet  unmatched prior detection at position $\hat{p} - \hat{D}_{\hat{p}}$ \cite{zhou2020tracking}. If the distance is more than the bounding box size of objects in the adjacent frames, a new tracklet is spawn. This association method uses displacement distance only.

\textbf{Objective Functions}
Focal loss is used as the training objective function to learn the heatmap \cite{lin2018focal,law2019cornernet}:

\begin{equation*}
\small
    L_{k} = \frac{1}{N}\sum_{xyc}\begin{cases}
      (1-\hat{Y}_{xyc})^{\alpha}\log(\hat{Y}_{xyc}) & \text{if $Y_{xyc} =1$},\\
      (1-Y_{xyc})^{\beta}(\hat{Y}_{xyc})^{\alpha}\log(1-\hat{Y}_{xyc}) &\text{otherwise}
    \end{cases}     
\end{equation*}
where $N$ is the number of objects, $Y_{xyc}$ a ground-truth heatmap corresponding to the annotated centers, and $\alpha =2 $ and $\beta=4$ are hyperparameters of the function. 

The regression objective functions for size and displacement use the L1 loss \cite{zhou2020tracking}:
\begin{equation}
    L_{size} = \frac{1}{N}\sum_{i=1}^{N} | \hat{S}_{\textbf{p}_{i}} - s_{i}|,
\end{equation}
where $s_{i}$ is the bounding box size of the $i$-th object at location $\textbf{p}_{i}$.
\begin{equation}
    L_{off} = \frac{1}{N}\sum_{i=1}^{N} |\hat{D}_{\textbf{p}_i^{(t)}} - (\textbf{p}_i^{(t-1)} - \textbf{p}_i^{(t)})|,
\end{equation} 
where $\textbf{p}_{i}^{(t)}$  and $\textbf{p}_{i}^{(t-1)}$ are tracked centers.


\section{Proposed Method}
Our proposed tracked object bounding box prediction (CenterTrack++) is built upon the CenterTrack tracking method in \cite{zhou2020tracking}. The original CenterTrack association method only uses single-frame tracked center offsets to associate objects through time, this method may fail in long-range tracklets or when occlusions occur as the occluded object's identity tends to assign to the object that occludes it. Under those cases, a single center displacement is not sufficient to obtain accurate tracking results. Therefore, additional size prediction is proposed to allow the tracking algorithm to better deal with the identity association by taking the overlapping of prior tracked object bounding boxes and predicted tracked bounding boxes into consideration. With two additional outputs on top of the existing tracking-conditioned CenterTrack method, the number of IDs can be easily reduced and tracking accuracy will then be improved, resulting in a better MOT tracker.

\subsection{Tracked Object Bounding Box and IOU Prediction (CenterTrack++)}
\begin{figure}[t]
  \centering
\includegraphics[width = 0.8\linewidth]{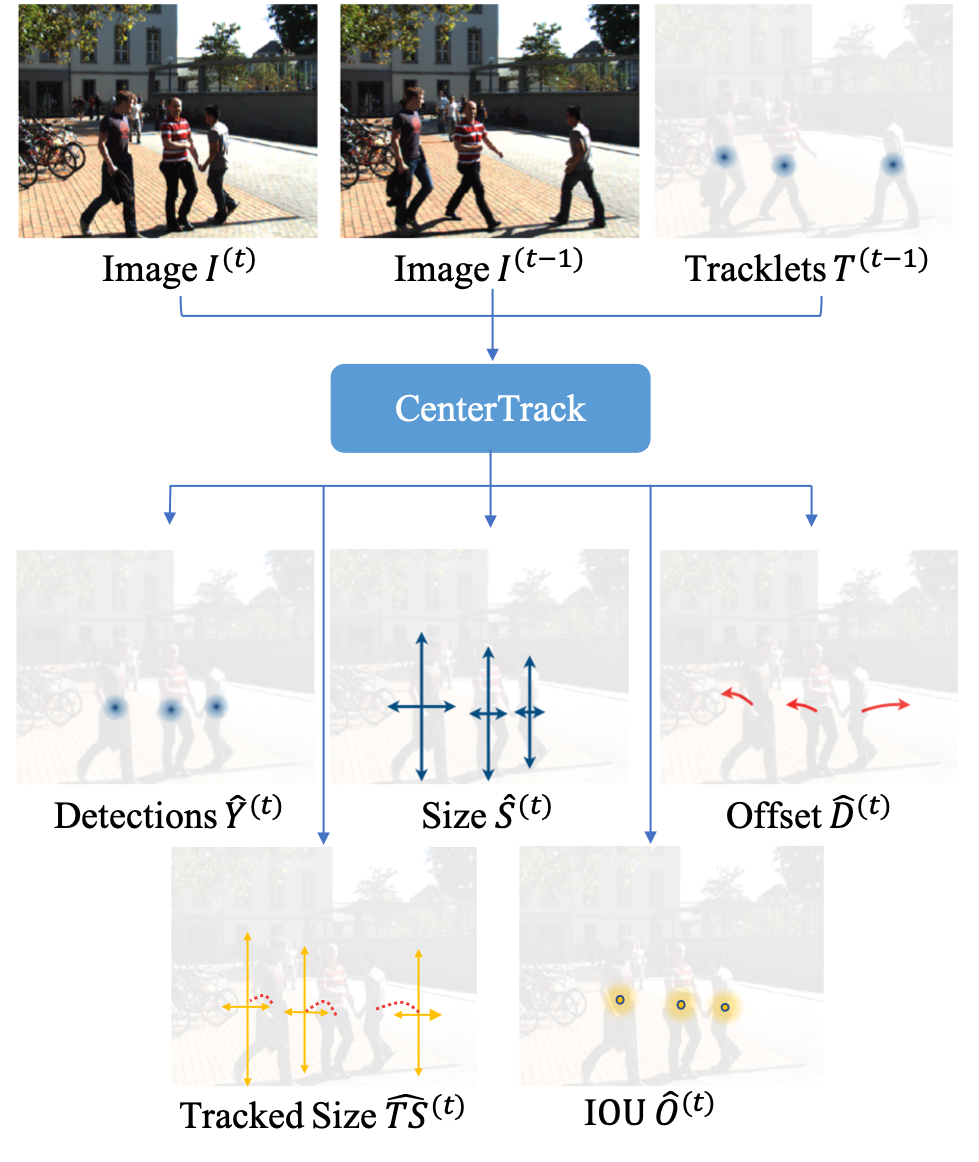}
\caption{Illustration of proposed CenterTrack++. Only two new output branches (Tracked Size and IOU) are added to the original CenterTrack \cite{zhou2020tracking} framework.}
\label{fig:ct+}
\end{figure}

Inspired by the idea of IOU distance in SORT \cite{sort} and IOU-Tracker \cite{iou_trackor}, IOU information is used in the association. To enable IOU distance calculation, prediction of tracked object bounding box in the prior frame with long-range tracklet lifetime is as shown in Figure \ref{fig:ct+}.

\textbf{Tracked Object Bounding Box Prediction.} There are two possible ways to predict tracked object bounding box in the previous frame based on the current frame.

\begin{itemize}
    \item \textbf{Tracking\_wh} Similar to the learning of center offset in the original CenterTrack model, the width and height difference $\hat{TS}\in \mathbb{R}^{\frac{W}{R}\times \frac{H}{R} \times 2}$ of the object's bounding box in the current frame and the previous frame is learned. This is used to predict the bounding box of the tracked object in the prior frame. For each detected object at location $\hat{\textbf{p}}$,  $\hat{TS}_{\hat{p}^{(t)}}^{(t)}$ is the difference of object size in the current frame $\hat{s}^{(t)} = (\hat{w}^{(t)},\hat{h}^{(t)})$ and previous frame $\hat{s}^{(t-1)}=(\hat{w}^{(t-1)},\hat{h}^{(t-1)})$, calculated by $\hat{s}^{(t)} - \hat{s}^{(t-1)}$, where $\hat{w}$ and $\hat{h}$ are width and height of object bounding box at location $\hat{\textbf{p}}$.
    
    \item \textbf{Tracking\_ltrb} Apart from learning width and height difference, we can also use offsets of the left, top, right, and bottom (ltrb) of the bounding box from the center in the prior frame instead, thus $\hat{TS}\in \mathbb{R}^{\frac{W}{R}\times \frac{H}{R} \times 4}$. For each detected object at location $\hat{\textbf{p}}$,  $\hat{TS}_{\hat{p}^{(t)}}^{(t)} = (\hat{x}^{(t-1)}-\frac{\hat{w}^{(t-1)}}{2},\hat{y}^{(t-1)}+\frac{\hat{h}^{(t-1)}}{2},\hat{x}^{(t-1)}+\frac{\hat{w}^{(t-1)}}{2},\hat{y}^{(t-1)}-\frac{\hat{h}^{(t-1)}}{2})$, $\hat{x}$ and $\hat{y}$ are horizontal and vertical coordinates of $\hat{\textbf{p}}$.
\end{itemize}

\textbf{IOU Prediction.} 
To further suppress inaccurate association, the IOU value of the bounding box of the same target in adjacent frames (IOU-adjacent) is learnt to provide a filtering threshold for unlikely associations. It is reasonable to assume that the IOU-adjacent is equal or smaller than IOU between detection in the prior frame and the regressed object bounding box based on the current frame (IOU-tracked). Any IOU-tracked that is smaller than the predicted IOU-adjacent should not be associated.
Therefore, we configure the network to learn the IOU-adjacent ($O\in [0,1]^{\frac{W}{R}\times \frac{H}{R} \times 1} $). For each detected object at location $\hat{\textbf{p}}$, $\hat{O}_{\hat{p}_{(t)}}^{(t)} = IOU( \hat{b^{(t-1)}} , \hat{b^{(t)})}$, where $\hat{b^{(t-1)}}$ and $\hat{b^{(t)}}$ are tracked bounding box in the previous and current frame at position $\hat{\textbf{p}}$ respectively. The model is conditioned to reason about how much overlapping of the same object's bounding box between adjacent frames. 

Therefore, two additional outputs would be produced by the model under this proposed method. The L1 loss objective function can be applied to them. 
For IOU prediction:
\begin{equation}
    L_{IOU} =  \frac{1}{N} \sum_{i=1}^{N}|\hat{O}_{\textbf{p}_{i}}^{(t)} - IOU(b^{(t-1)}_{\textbf{p}_{i}},b^{(t)}_{\textbf{p}_{i}})|,
\end{equation}
where $b^{(t-1)}_{\textbf{p}_{i}}$ and and $b^{(t)}_{\textbf{p}_{i}}$ are tracked ground-truth bounding boxes.

In the case of Tracking\_wh approach for tracked object bounding box prediction:
\begin{equation}
     L_{tracked\_size} = \frac{1}{N} \sum_{i=1}^{N} | \hat{TS}_{\textbf{p}_{i}}^{(t)} - (s_{i}^{(t-1)} - s_{i}^{(t)})|.
\end{equation}

In the case of Tracking\_ltrb approach for tracked object bounding box prediction:
\begin{equation}
    \begin{split} L_{tracked\_size} =
    \frac{1}{N} \sum_{i=1}^{N} |\hat{TS}_{\textbf{p}_{i}}^{(t)} - ( & x_{i}^{(t-1)}-\frac{w_{i}^{(t-1)}}{2},\\
 & y_{i}^{(t-1)}+\frac{h_{i}^{(t-1)}}{2}, \\
 & x_{i}^{(t-1)}+\frac{w_{i}^{(t-1)}}{2}, \\
 & y_{i}^{(t-1)}-\frac{h_{i}^{(t-1)}}{2})|.
\end{split}
\end{equation}

\subsection{Association Steps}
 As explained in \cite{sort}, IOU distance can implicitly handle short-term occlusions caused by passing targets. Since we can predict the tracked object bounding boxes from the current frame, IOU distance cost matrix can be naturally incorporated into the association process. IOU distance cost is calculated by $1 - IOU(\hat{b}^{(t-1)},\hat{tb}^{(t)})$, where $\hat{b}^{(t-1)}$ is detected bounding box in the prior frame and $\hat{tb}^{(t)}$ is tracked bounding box prediction from the current frame. If the IOU distance cost is more than $\hat{O}^{(t)}_{\hat{p}_{i}^{(t)}}$ at the detected point ${\hat{p}_{i}^{(t)}}$, we set the corresponding cost to infinity which effectively prevents from unlikely associations. After distance cost matrix computation, a simple greedy matching algorithm is employed to assign object identities.

As displacement and IOU distance cost matrices are now available, we further explored the possibilities of using different combinations and orders of matrices during association. Not only using a single matrix, two matrices can be summed up together to produce a combined matrix for the association. Additionally, we can use two matrices sequentially. Specifically, after one round of a simple greedy matching with one distance cost matrix, a different cost matrix can be used to associate the remaining unmatched detections and tracklets further with another round of greedy matching. Tracking performances of different association methods are reported in section 4 under Ablative Studies.
   
\section{Experiments}
\subsection{Dataset and Metrics}
We use MOT17 \cite{milan2016mot16} dataset to train and evaluate the proposed method in our paper. MOT17 contains 7 sequences for training and test respectively. The videos were captured by stationary cameras mounted in high-density scenes with heavy occlusion. Only pedestrians are annotated and evaluated. The video framerate is 25-30 FPS. Since MOT17 does not provide an official validation split, we split each training sequence into halves, the first half for training and the second one for validation in our ablative studies. Our main results are reported on the test set. We followed the same metric used in the original CenterTrack model, the CLEAR metric \cite{Bernardin2008} and Identification $F_1$ score (IDF1) \cite{ristani2016performance}, to evaluate overall tracking accuracy.

\subsection{Implementation Details}
Our implementation is based on CenterTrack, DLA \cite{yu2019deep} is used as the network backbone with Adam optimizer \cite{kingma2017adam} at a learning rate of $1.25e -4$ and a batch size of 16. We use standard data augmentations include horizontal flipping, random resized cropping, and color jittering. For all experiments, we use 70 epochs for the network training. 
The learning rate drops by $1/10$ at the 60th epoch. We train and test the proposed model with three GTX 1080 Ti GPUs.

The input size is resized to $960 \times 544$, with downsampling $R =4$. Followed the recommended parameters in CenterTrack, we also use random false positive ratio $\lambda_{fp} =0.1$ and random false negative ratio $\lambda_{fp} =0.4$ to generate noises in the dataset to train a robust tracking-conditioned object detector. Similarly, we only output tracklets with confidence of  $\theta = 0.4 $ and above and render heatmap with a threshold $\tau = 0.5$. The network is pre-trained on CrowdHuman dataset \cite{shao2018crowdhuman} before training on MOT17 dataset. However, unlike original CenterTrack, we use long-range tracklets, tracklet lifetime = 30, discarding the unmatched tracklets only after 30 frames.

\subsection{Ablative Studies}
As described in the previous section, we experimented cost matrix with different combinations and orders:
\begin{enumerate*}
    \item Displacement only (DIS), used in original CenterTrack;
    \item IOU only (IOU);
    \item IOU and displacement (Combined);
    \item IOU first followed by displacement (IOU$\to$DIS);
    \item Displace first followed by IOU (DIS$\to$IOU).
\end{enumerate*}

\begin{table}[]
\begin{center}
\caption{Results on MOT17 validation set using the tracking\_wh approach.}

\par\smallskip\label{tab:tracking_wh}
\small
\begin{tabular}{|c|c|c|c|c|c|}
\hline
\textbf{Association}  & \textbf{IDF1$\uparrow$} & \textbf{MOTA$\uparrow$} & \textbf{IDs$\downarrow$} & \textbf{FP$\downarrow$} & \textbf{FN$\downarrow$} \\ \hline
DIS                  & 69.2          & 66.2          & 219          & 3.9         & 29.5      \\
IOU                           & \textbf{71.1} & 66.7         & 204           & \textbf{3.6} & 29.3          \\
Combined                      & 70.9         & 66.2         & 233           & 3.9          & 29.6        \\
DIS$\to$IOU         & 70.0        & 66.2        & 218           & 3.9          & 29.5          \\
IOU$\to$DIS         & 69.8         & \textbf{66.8} & \textbf{185}  & \textbf{3.6} & \textbf{29.2} \\ \hline
\end{tabular}
\end{center}
\vspace{-5mm}
\end{table}

\begin{table}[]
\begin{center}
\caption{Results on MOT17 validation set using the tracking\_ltrb approach.}
\par\smallskip\label{tab:tracking_ltrb}
\small
\begin{tabular}{|c|c|c|c|c|c|}
\hline
Association & IDF1$\uparrow$ & MOTA$\uparrow$ &IDs$\downarrow$ & FP$\downarrow$ & FN$\downarrow$ \\ \hline
DIS                  & 69.2          & 66.2          & 219          & 3.9         & 29.5      \\
IOU                   & \textbf{72.4}     & \textbf{66.7}     & 191           & 3.8             & 29.2            \\
Combined              & 70.8              & 66.5              & 236           & 3.8             & 29.3            \\
DIS$\to$IOU & 70.5              & 66.6              & 202           & 3.8             & 29.2            \\
IOU$\to$DIS & 71.4              & \textbf{66.7}     & \textbf{166}  & 3.8             & 29.2            \\ \hline
\end{tabular}
\end{center}
\vspace{-5mm}
\end{table}

The result of IOU only association method in Table \ref{tab:tracking_wh} and \ref{tab:tracking_ltrb} confirm the benefits of using additional tracked object bounding box prediction to reduce IDs of object tracking, both IDF1 improves compared to the baseline DIS CenterTrack tracking algorithm, with a significant 3.2\% IDF1 improvement from baseline method using tracking\_ltrb approach in Table \ref{tab:tracking_ltrb}. Since our proposed method focuses on improving CenterTrack's association ability, not detection capability, a small improvement in MOTA is expected and observed as MOTA metrics MOTA penalizes detection errors and IDs while IDF1 focuses on the tracking accuracy of detected objects. \cite{DBLP:journals/corr/RistaniSZCT16}.

Comparing the overall performance between tracking\_wh and tracking\_ltrb approach to predict tracked object bounding box, it is observed that the use of ltrb offsets is more effective to regress the width and height of the bounding box of the tracked centers from the current detected centers compared to just learning from the size offset between adjacent frames.

However, the idea of sequential matching using different matrices yields little or no improvement in association accuracy compared to single IOU matching, implying that single IOU matching is sufficient to provide accurate tracking results. Additionally, the idea of a combined distance matrix does not necessarily improve tracking accuracy neither, this could be due to the addition of two distance matrices with different scales, which can be considered as noises to corrupt the associating power of the other matrix.

\begin{figure}[!tbh]
\begin{minipage}[]{1\linewidth}
  \centering
\includegraphics[width=0.9\linewidth]{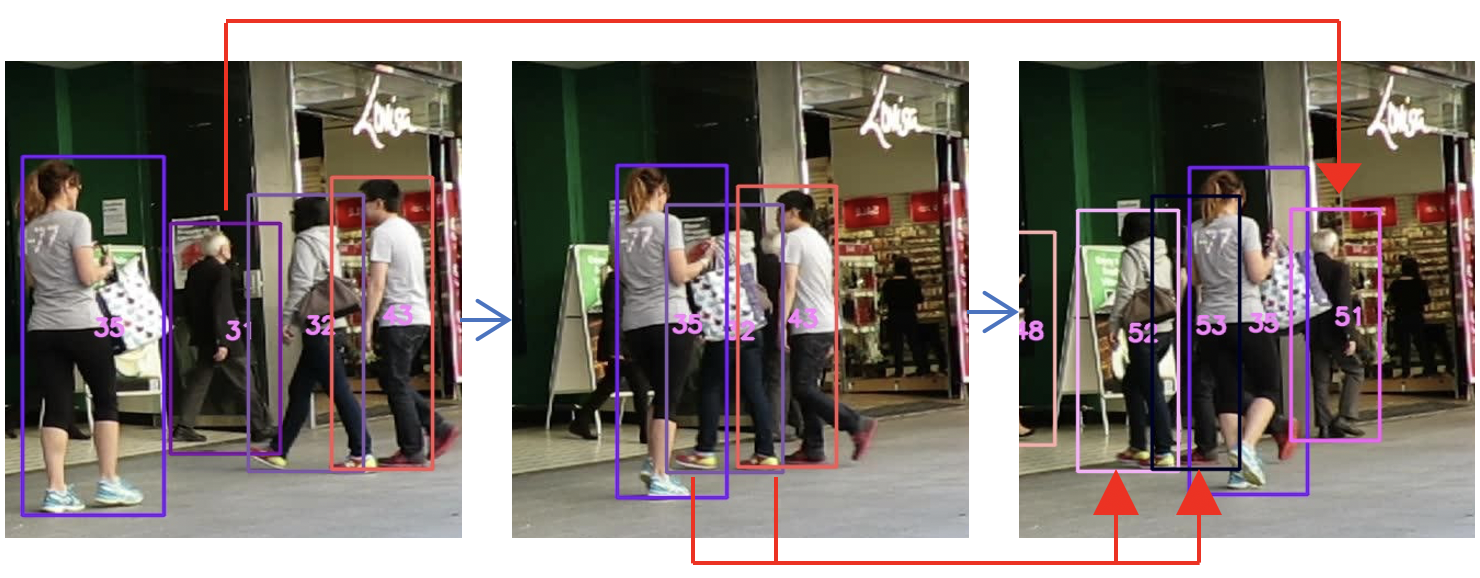}
  \centerline{(a) CenterTrack}\medskip
\end{minipage}
\begin{minipage}[]{1\linewidth}
  \centering
\includegraphics[width=0.9\linewidth]{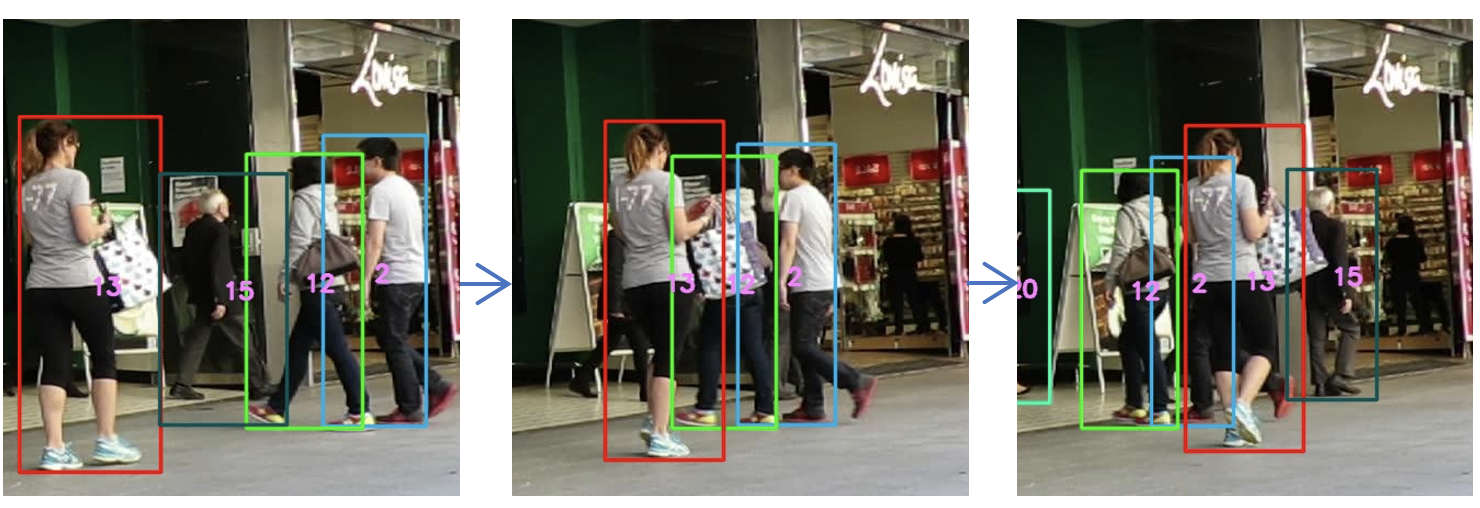}
  \centerline{(b) CenterTrack++}\medskip
\end{minipage}
\caption{Comparison of the tracking results on the sequence "MOT17-09" validation when short-term occlusions occur with each arrow representing 10-frame interval.}
\label{fig:occlusion}
\end{figure}

\begin{figure}[htp]
\centering
\begin{minipage}[]{0.4\linewidth}
\centering
\includegraphics[width=0.8\linewidth]{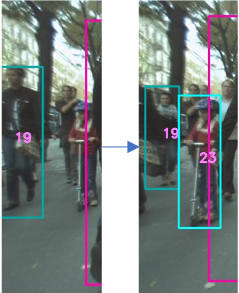}
  \centerline{(a) CenterTrack}\medskip
\end{minipage}
\begin{minipage}[]{0.4\linewidth}
\centering
\includegraphics[width=0.8\linewidth]{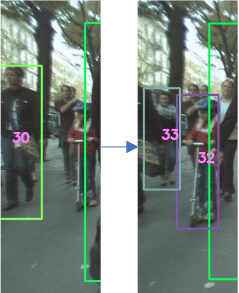}
  \centerline{(b) CenterTrack++}\medskip
\end{minipage}
\caption{Comparison of the tracking results on the sequence "MOT17-02" validation set when the leftmost person exited the frame. CenterTrack assigns the same ID to different person(ID:19), while CenterTrack++ does not.}
\label{fig:exit}
\end{figure}

CenterTrack++ is more robust in object tracking compared to CenterTrack in cases when occlusions occur or objects exit the frame. Figure \ref{fig:occlusion} and \ref{fig:exit} demonstrate CenterTrack++'s ability to track objects accurately during those cases while CenterTrack fails.


\subsection{Test Result}
From the ablative studies, we found out that the use of IOU distance in the association step under the tracking\_ltrb approach yields the best performance. We adopt the best method and evaluate its performance on MOT17 test data. The results are shown in Table \ref{tab:test}. It is shown that our method can reduce the IDs significantly by 22.6\% and obtain a notable improvement of 1.5\% in IDF1 score compared to the Original CenterTrack under the same tracklet lifetime. Our method obtains the best performance based on MOTA, IDF1 and IDs evaluation metric among trackers only using spatial features for association. Compared with FairMOT \cite{zhang2020fair} that employs re-identification based on additional appearance features, our method still obtains better IDs score (2352 vs. 3303).

\begin{table}[!tbh]
\centering
   \caption{Comparison of the state-of-the-art methods under "private detector" protocol. Note: S=Spatial, A=Appearance.}
     \label{tab:test}
\small
\begin{tabular}{|c|c|c|c|c|}
\hline
\text Tracker  & \begin{tabular}[c]{@{}c@{}}Association \\ Features\end{tabular} & MOTA$\uparrow$ & IDF1$\uparrow$ & IDs$\downarrow$ \\ \hline
TubeTK\cite{pang2020tubetk}              & S                       & 63            & 58.6          & 4137         \\
CenterTrack\cite{zhou2020tracking}    & S                       & 67.8          & 64.7          & 3039         \\
Ours          & S                       & \textbf{68.1}          & \textbf{66.2}          & \textbf{2352}         \\\hline
SST\cite{deepaffinity}               & A                    & 52.4          & 49.5          & 8431         \\
CTrackerV1\cite{peng2020chainedtracker}        & S+A                      & 66.6          & 57.4          & 5529         \\
DEFT\cite{Chaabane2021deft}	 &S+A	&66.6	&65.4	& \textbf{2823} \\
FairMOT\cite{zhang2020fair}         & S+A                    & \textbf{73.7}          & \textbf{72.3}          & 3303   \\\hline
\end{tabular}
\end{table}

\section{Conclusion}
In this paper, we propose tracked object bounding box and overlapping prediction outputs onto the CenterTrack tracking algorithm, which reduces the IDs and improves overall tracking accuracy. The extra prior tracked object bounding box and overlapping prediction enable the use of the IOU distance matrix to associate objects across frames more accurately. Experiments on MOT17 test dataset under private protocol show that our proposed method achieves the best performance among the trackers only using spatial features in the association.

\bibliographystyle{IEEEbib}
\bibliography{icme2021template}

\end{document}